\documentclass[graybox]{svmult}

\usepackage{mathptmx}       
\usepackage{helvet}         
\usepackage{courier}        
\usepackage{type1cm}        
\usepackage{makeidx}         
\usepackage{graphicx}        
\usepackage{multicol}        
\usepackage[bottom]{footmisc}

\makeindex             

\begin{document}

\title*{A Compendium on Network and Host based Intrusion
Detection Systems}
\author{Rahul-Vigneswaran K         \and
        Prabaharan Poornachandran   \and
        Soman KP}
\institute{Rahul Vigneswaran K \at
              Department of Mechanical Engineering, Amrita   Vishwa Vidyapeetham, Amritapuri, India  \\
              Phone: +91-8667527705\\
              \email{rahulvigneswaran@gmail.com}           
           \and
           Soman KP \at
              Center for Computational Engineering and Networking (CEN), \\Amrita School of Engineering, Amrita Vishwa Vidyapeetham, Coimbatore, India  
            \and
           Prabaharan Poornachandran \at
              Center for Cyber Security Systems and Networks, Amrita School of Engineering, \\Amrita Vishwa Vidyapeetham, Amritapuri, India. \\  
}
%
%
\maketitle

\abstract*{The techniques of deep learning have become the state of the art methodology for executing complicated tasks from various domains of computer vision, natural language processing, and several other areas. Due to its rapid development and promising benchmarks in those fields, researchers started experimenting with this technique to perform in the area of, especially in intrusion detection related tasks. Deep learning is a subset and a natural extension of classical Machine learning and an evolved model of neural networks. This paper contemplates and discusses all the methodologies related to the leading edge Deep learning and Neural network models purposing to the arena of Intrusion Detection Systems.}

\abstract{The techniques of deep learning have become the state of the art methodology for executing complicated tasks from various domains of computer vision, natural language processing, and several other areas. Due to its rapid development and promising benchmarks in those fields, researchers started experimenting with this technique to perform in the area of, especially in intrusion detection related tasks. Deep learning is a subset and a natural extension of classical Machine learning and an evolved model of neural networks. This paper contemplates and discusses all the methodologies related to the leading edge Deep learning and Neural network models purposing to the arena of Intrusion Detection Systems.}

\section{Introduction}
\label{intro}
The field of cybersecurity has recently gained a significant importance due to the digitalization of developing countries. In several other perspectives, it is even considered as a new way of warfare, due to which several nations are pushing towards a more robust firewall for all the nation's digital assets. This lead to the development of Intrusion Detection System (IDS), which acts as the first layer of the system and helps in identifying the inbound threats which must be apprehended for safety reasons. 

IDSs are a strategically placed layer in the system that detects threats and monitors the exchanged packets. These type if security layers are there in existence since 1980's \cite{1}. They use the hard-coded algorithms to analyze the patterns in the captured data and filter the anomalies. These have proven effective in certain cases but due to its inherent if-else nature, the problem of false positive or false negative alarms was a common issue. Due to the false triggers of alarms by the IDS, network administrators faced increased difficulty in handling intrusion reports. Also, their inability to tackle unknown attack types rendered them useless. 

Commonly there are two types of IDS that exists, namely : 
\begin{itemize}
    \item \textbf{Network based Intrusion Detection System (NIDS):}  
    NIDS is a platform which is independent and aims at detecting intrusions by examination of network traffic and monitoring multiple hosts. These type of system access traffic by behaving like a parasite to the network hub, a switch specifically designed for port mirroring. For accuracy and efficiency, the sensors are situated at the strategically chosen choke locations, often in the demilitarized zone. These receptors collect all the network data and analyze the contents of packets. 
    \item \textbf{Host-based Intrusion Detection System (HIDS):} It usually has a singular agent embedded onto a host system that detects intrusions by going through the logs, calls made by the system, modifications of file-system and other states and activities. In HIDS, a software agent is a sensor. 
\end{itemize}
Traditionally, these IDS systems take advantage and make use of the signatures of known attacks to identify the inbound attacks. But due to the rapid advancement of new types of malware whose signatures are unknown, the filtering of ever-increasing types of state of the art malware becomes difficult. This paved way for the conjunction of machine learning and cybersecurity. IDS in machine learning is a classification problem. Therefore, possibly an unbiased dataset is used for training the model and that model is optimized for integrating into the IDS to predict malware more accurately. This whole prediction principle eliminates the complete dependency of malware signatures for detection, hence acting as an advantage over traditional IDS. 

For the prediction model to work more efficiently, we need to feed it with a lot of training data. Often these raw data are unlabelled and come with improper feature engineering. The pitfall with traditional machine learning is that the training data must be labelled properly for the model to discover patterns and therefore provide prediction results. But this is not always possible to the vast size of the data. This is where Deep Learning (DL) has explicit advantages over conventional algorithms in scenarios of processing data that is unstructured. Pattern discovery can be done without the ancillary task of labelling. In classical machine learning theory, engineering features become an imperative part. DL lacks the feature scrutiny. This becomes vital in plots where the objective is to segregate fraudulent sequences versus non-fraudulent sequences and justify how we reached our conclusions. Fortunately, the ultimate motive of an IDS is this, which makes deep learning a perfect match for Intrusion Detection Systems. 

\section{Big Data Analytics}
\label{sec:1}
Presently, the current networks of computers are generating a huge amount of data traffic in a fast past. This generation of data is increasing continuously and the types of data that the network traffic contains is also different. Current traffic of the public satisfies all the three V's of big data namely, volume, velocity, and variety \cite{2}. Big Data is a term implicates the large volume of data (structured \& unstructured) that deluge business worldwide at a faster rate than anyone can predict. A large pool of data doesn't always signify that everything is useful and this calls for a tool that can extract the essence from it to decide the cardinal step. This is where the Analytics of the so-called Big Data comes into play. Rapid digitization has to lead the current world to a point where we might run out storage capabilities soon. This implicates a much-required need to deduce the pool of data into useful decisions. 

Detection of intrusion often involves analyzing Big data. This is defined as a research problem which the technologies of computing which are mainstream cannot cope up with the flow of information. Even an isolated source of the event can lead to Big data challenges. According to \cite{3}, a 1 Gigabytes/second of continuous network data can cause issues related to Big Data for Intrusion Detection even when inspecting deep data packets. Traditional storage methods like Relational databases inherently have the capacity to effectively gamut against the nature of Big data challenges caused by detection of intrusion. A common solution used to handle such problems is an open sourced tool called Hadoop. It is a storage platform which is distributed and is generously open-sourced. It has the capability to run on most of the hardware and has been exploited to better entertain the Big Data storage requirements of gigantic Volume and agility along with potentially very versatile data structures that are natively heterogeneous. Hadoop is collective name for several other technologies such as Pig, MapReduce, Hive, HDFS, etc.. \cite{4} proposes the use of these technologies to overcome the hurdles in big data and states that the 3Vs are not sufficient to handle these issues and hence introduces the 3Cs concept – Cardinality, Continuity, and Complexity. 

The process of mining this Big data and analyzing it is called Big Data analytics. The fast-paced emendations in aspects of processing, storing and analyzing of Big Data in recent years includes :
\begin{itemize}
    \item The rapid decrease in storage expense and CPU power.
    \item Cost-effectiveness and Flexibility of Cloud computing and Datacenters in accordance with storage and elastic computation.
    \item Development of revolutionary and new frameworks such as Hadoop.
\end{itemize}
The ultimate objective of Big Data analytics for IDS is to ensure promising, market-ready real-time intelligence. Even though it has a compelling future, there are several problems (such as data provenance, privacy) that must be diminished in order to unleash the actual potential of it.  

\section{Neural Networks}
\label{sec:2}
A Neural Network is a paradigm related to the processing of information whose thought process is motivated by the way the nervous systems of several biological parts, such as the brain, analyze and segregate information. The important aspect of this approach is the novel assembly of the system that processes information. It is built with numerous neurons which are highly connected with each other, working in unison to understand the training dataset and solve problems specific to the set. These, like people, learn by example. Learning in biological environment systems involves tuning of these links that between the neurons of multiple layers. This is true of artificial neural networks as well.

\section{Deep Learning}
\label{sec:3}
It is a natural extension to the existing field of ML exercised with models that are motivated by the assembly and function of brain cells. One of the key aspects of Neural networks is scalability which signifies that the result's superiority is directly proportional to the amount of data that we use but it also implies the need for significantly more computational time. In addition to the advantage of easy scaling, another often stated plus point is their intuitive ability to eliminate the burden of large-scale feature engineering. Especially, DL shines in scenarios which require analogue input-outputs which enables it to work with RGB pixel data from images, text data from documents or audio data from music files rather than limiting the user to use quantities in a tabular format.

\section{Neural Networks and Deep learning techniques for Network based IDS (NIDS)}
\label{sec:4}

The aim of an IDS is to detect abnormalities in the traffic and mitigate a possible intrusion. This must be done in an efficient way considering the ever-increasing variants of malware and types of attacks in the field of cybersecurity. The use of signature-based and traditional if-else algorithm based detection layers have been drawn ineffective which calls for the inclusion of AI in the field of NIDS to predict an inbound attack that is likely to cause damage or theft of digital assets. 

Many researchers have been actively working for years now to experiment with different types of techniques and datasets to establish a Network-based IDS that has superhuman accuracies and capabilities. The use of deep learning and artificial neural networks have proven effective in such applications due to the improvement in the computational power of the commercially viable processors. To overcome the principle challenges of a traditional intrusion detection systems, \cite{5} has proposed and implemented an analysis method that combines the abilities of kernel PCA and LSTM-RNN. To obtain the desired result, features like pre-processing of data, extraction of features, detection of attacks have been incorporated into the layer proposed by it. They have used the NSL-KDD dataset for benchmarking and has outperformed the capabilities shown by SVM, neural network, and Bayesian methods. 

The experiments done by \cite{6} with LSTM-RNNs on the KDD99 dataset has shown an accuracy of 93.82\% due to their inherent ability to correlate consecutive records of connections by peeking back in time. An avant-garde DL technique for NIDs has been discussed by \cite{7}, in which they have adduced an NDAE (Non-symmetric Deep Auto-Encoder) for learning in unsupervised conditions. The outcome of it when benchmarked against both KDD99 and NSL-KDD datasets has shown a 5\% improvement in accuracy and 98.81\% reduction in training time. \cite{8} has done a comparative study of deep learning like vanilla DNN,  self-taught learning (STL) approach, RNN based LSTM on the KDD99 dataset and then compared with a baseline shallow algorithm that uses multinomial logistic regression to evaluate if deep learning models perform better in this scenario. An important observation that STL model could be used in a scenario where the data is unclean is made by this experimentation.  

As a whole, when various methods of deep learning and neural network methods are compared with each other and studied, one can attain a state of understanding that different scenarios require different AI models. Even in NIDS, most of the models can be made to have high accuracy when proper hyperparameter tuning is done. One of the major drawback reported by all the researchers is the availability of latest ID dataset. All the research that is currently done is based on either original or modified old datasets which don't consider the more advanced threats that the modern day NIDS is required to filter.  

\cite{11} takes a rather different approach to the whole intrusion detection system.  The model learns the user behaviour and when it the new behaviour doesn't match it,  the alarm is triggered indicating a possible security breach. In this, a backpropagation neural network has been used and the results turned out with a 96\% accuracy in detection of events that are unusual, with a false-alarm rate of 7\%. On the other hand, \cite{12} takes into account a variety of neural network structures to figure out the appropriate model for the current scenario. Fortunately, an early barring corroboration method is also administered during the training to substantially escalate the capability of the network to generalize, which lends an accuracy rate of 91\% and 87\% with two hidden neuron layers and one hidden neuron layers respectively.  

The use of Genetic Network Programming is undertaken by \cite{13} and has been combined with the fuzzy set theory for better results. KDD99 and 98 datasets have been used for benchmarking which lends success rates that are high compared with other ML techniques and data mining which is crisp with GNP. A common problem faced by all the existing approaches are false-alarm rates, however small it may be, they can cause a lot of unwanted distractions to the system administrators and an ideal IDS would have zero false-alarm rates. In the path towards the ideal IDS, \cite{14} has tried to enhance the system's performance by using keyword selection and neural networks. \cite{15} have applied the technique of deletion of features one at a time for performing experiments on NNs and SVMs to estimate the hierarchy of priority of DARPA ID dataset's input features. Hence, doing so, they have identified the important features for IDS and as a result, drastically reduced the training time. 

Taking into account the reality that, terabytes of data pour through the IDS if we tried to analyze the entire traffic flow, \cite{16} has developed an anomaly detector that considers the only certain portion of attack space that is deemed vital for the detecting things such as protocols, that are forecastable. Within the limitations of the experimental setup, the conclusions show that this model truly responds to the anomalies, not to signatures of known attacks. 

\cite{17} has taken a notable approach to IDS. Convolutional neural networks have a structure that favours computer vision and has provided significant results in that field. This significance is being transported to cybersecurity by using CNN-RNN in IDS layers and one of the main reason for choosing such an approach is the model's ability to extract high-level features and represent the consolidated form of low-level features. Both \cite{18} and \cite{19} have used KDD99 as their benchmarking datasets and the former tries to run a comparison between shallow and deep neural networks in the scenario of NIDS and approves that many of the inherent characteristics of deep networks provide an overall advantage for it compared to shallow networks while the later experiments the usage of stacked-RNNs. The stacked-RNN models show reduced false-alarm rate due to its nature and potential to learn complex temporal behaviours quickly. 

\section{Neural Networks and Deep learning techniques for Host based IDS (HIDS)}
\label{sec:6}
A host-based intrusion detection system (HIDS) is a scheme that supervises and logs activities of the computer on which it is installed to alert the admin in case the protocols are breached. Host-based IDS can be conceived as an agent that oversees and analyzes whether anything or anyone (both internal and external), has bypassed the security policy of the system. NIDS is usually placed at the demilitarized zone of the network. Its purpose is to check the incoming data and eliminate threats and then pass the data forward to the proprietor if the existence of potentially malicious packets is non-existent with the transmission of data. It is more focused on the singular local node, unlike NIDS. \cite{9} has made comprehensive experimentation by using DNNs and SVMs (Support Vector Machines) on the KDD-DARPA benchmarking dataset for making an Host-based IDS to overcome the drawbacks of the traditional hard-coded logic and algorithms. The accuracy has been concluded as greater than 99\% with a training time of 17.77 sec. But here too the pitfall of inability to train with new attack patterns have been cautioned.  

An approach where Radial basis functions neural networks as containers are being used by \cite{10} for host-based IDS which proves robust compared to the soft computing methods with very low training time and high accuracy in prediction. Therefore the answer to the question about superiority is very important. The ultimate solution is using both NIDS and HIDS in conjunction with each other and AI integrated into them. In order for the host (and network) to remain secure, it is essential that we should select the right IDS scheme that is appropriate for the provided circumstances.

\section{Conclusion}
\label{sec:7}

 As the importance of IDS in the day-to-day digital world is more obvious, the necessity for a more accurate model with no false alarm rate has become apparent. To attain an algorithm that is more robust to the ever-increasing variety of malware and threats, researchers have taken an array of paths and have succeeded in achieving exponentially high accuracy rates while benchmarking with the existing publicly available dataset. It is clear that, with proper hyperparameter tuning, most of the model existing in the market can achieve superhuman ability in identifying inbound threats, which leads us to the conjunction of most of the research papers that have been made on artificial intelligence based IDS. Due to the basic principle of deep learning which dictates that more data equals more accuracy, we need to train the models with a dataset that has signatures and features of more recent advanced cyber warfare attacks, in order to prepare the research stage IDS into a robust market-ready product. 
 
\begin{acknowledgement}
This paper would have not been possible without the continued guidance of Dr. KP. Soman of Centre for Computational Engineering and Networking (CEN), Amrita Vishwa Vidyapeetham, India and Dr. Prabharan Poornachandran of Centre for Cyber Security Systems and Networks, Amrita School of Engineering, Amrita Vishwa Vidyapeetham, Amritapuri, India. Also, We thank the anonymous reviewers for their insightful comments and suggestions.
\end{acknowledgement}

\end{document}